# Rotated Feature Network for multi-orientation object detection

Zhixin Zhang, Xudong Chen, Jie Liu, Kaibo Zhou*

Huazhong University of Science and Technology

{zhixinzhang, xudchen, jie_liu, zhoukb}@hust.edu.cn

## Abstract

*General detectors follow the pipeline that feature maps extracted from ConvNets are shared between classification and regression tasks. However, there exists obvious conflicting requirements in multi-orientation object detection that classification is insensitive to orientations, while regression is quite sensitive. To address this issue, we provide an Encoder-Decoder architecture, called Rotated Feature Network (RFN), which produces rotation-sensitive feature maps (RS) for regression and rotation-invariant feature maps (RI) for classification. Specifically, the Encoder unit assigns weights for rotated feature maps. The Decoder unit extracts RS and RI by performing resuming operator on rotated and reweighed feature maps, respectively. To make the rotation-invariant characteristics more reliable, we adopt a metric to quantitatively evaluate the rotation-invariance by adding a constrain item in the loss, yielding a promising detection performance. Compared with the state-of-the-art methods, our method can achieve significant improvement on NWPU VHR-10 and RSOD datasets. We further evaluate the RFN on the scene classification in remote sensing images and object detection in natural images, demonstrating its good generalization ability. The proposed RFN can be integrated into an existing framework, leading to great performance with only a slight increase in model complexity.*

## 1. Introduction

Object detection has drawn great research interests, driven by many important real-world applications, such as face detection [10, 29], crowded people counting [28, 30] and text detection [1, 13]. Over the past few years, although significant progress has been made in previous works [7, 20, 25, 34], object detection still remains a big challenge as object orientation can vary a lot, especially in remote sensing images. In this paper, we focus on the problem of multi-orientation object detection.

Owing to the perfect feature representation ability of deep ConvNets, deep learning-based methods, such as SSD [20], Faster R-CNN [25] and YOLO [24], have achieved great performance for general object detection in recent years. General detectors mainly focus on object detection in natural images, and are not appropriate for detecting objects with arbitrary orientations. Objects in remote sensing images are typical examples with multi-orientation. General detectors directly used in such images may not be effective for multi-orientation object detection, although methods like Faster R-CNN have achieved state-of-the-art performance.

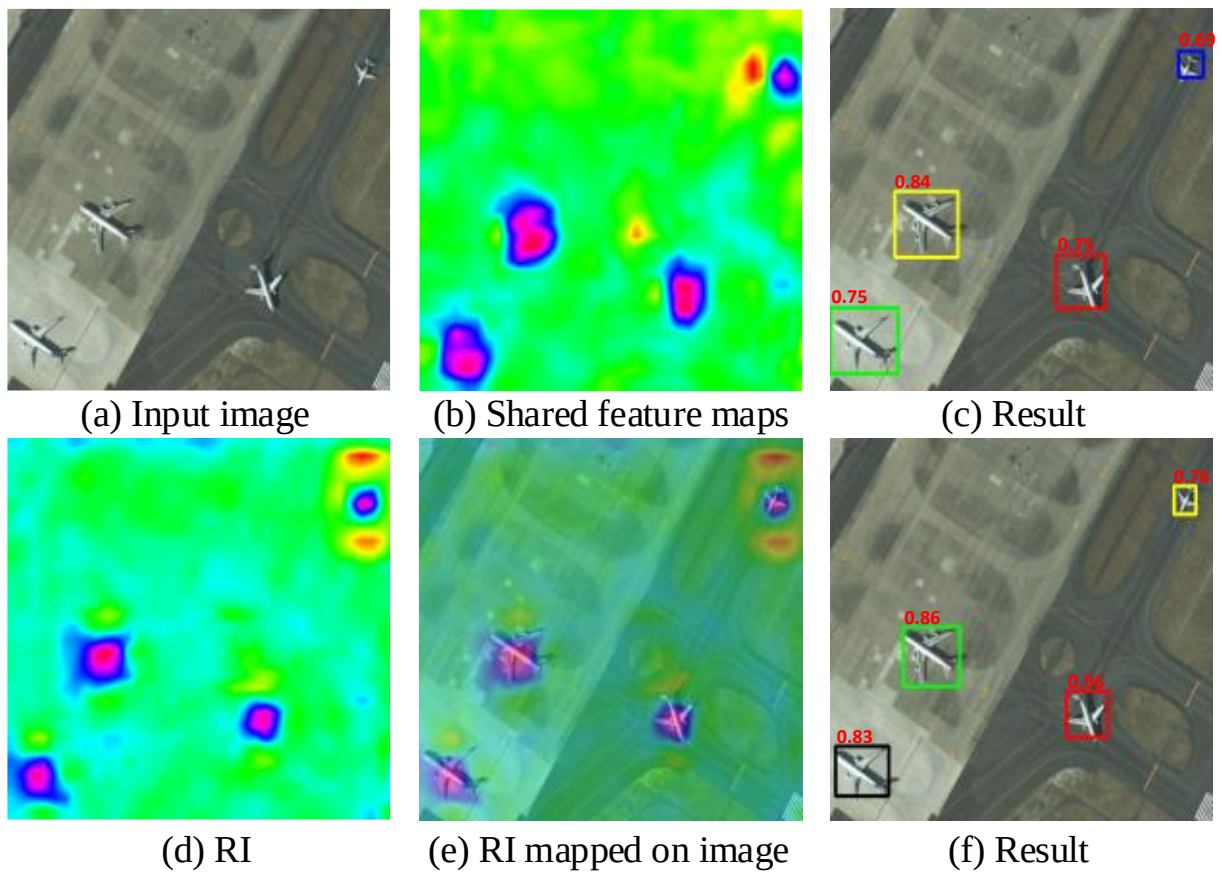

Figure 1. Visualization of results for the general detector and our method. Objects are localized with boxes in different colors. Values above boxes denote confidence scores. (a): input image; (b) response map of shared feature maps; (c) result by using general detector; (d): response map of rotation-invariant feature maps (RI); (e): RI mapped on image; (f): result by using RFN.

Multi-orientation objects are commonly seen in remote sensing and scene text images. Recently, some researchers have focused on handling object detection in such images. He et al. [13] firstly proposed a deep direct regression method for multi-orientation text detection, which performs regression by predicting offsets from a given point. Cheng et al. [2] presented an effective method to learn rotation-invariant features, which has been successfully applied where object orientations vary a lot, especially in remote sensing images. All the methods follow the pipeline: extracted feature maps of each proposal are fed into two tasks, i.e. classification and regression. A typical characteristic is that both tasks share the same feature maps, which are extracted from a deep ConvNet. However, there exists conflicting requirements between classification and regression tasks. Specifically, for classification task, it is expected that object of arbitrary orientations should be classified to the same class. It means that the classification task is insensitive to orientations. On the contrary, for regression task, the localization of an object is quite sensitive to orientations. Specially, the conflict is quite serious in multi-orientation object detection, especially in remote sensing images. Moreover, previous works like ORN [39] and RRD [17] have provided rotation-invariant features with qualitative or intuitive analysis, which cannot promise the reliability of rotation-invariant features.

In this paper, to address the above issues, we propose a novel and effective method for multi-orientation object detection, which is called Rotated Feature Network (RFN). Our method is based on an existing object detection method, i.e. Faster R-CNN, and extends the RoI pooling layer to the proposed RFN, which consists of two cascade components: Encoder unit and Decoder unit. The Encoder unit, partly inspired by SE block [14], is designed to reweigh the importance of rotated feature maps, and produces highly representational feature maps. In the Decoder unit, two types of feature maps are produced to meet the different requirements, i.e. rotation-invariant feature maps (RI) for classification and rotation-sensitive feature maps (RS) for regression. Besides, to make the rotation-invariance more reliable, we adopt a metric to quantitatively evaluate it, which is implemented by imposing a loss on RI. The comparison between a general detector (Faster R-CNN) and our method is visualized in Fig. 1. It can be seen that the RI is more insensitive to orientations and helpful to achieve better classification performance, and the RS can facilitate to generate more accurate boxes, compared with the shared feature maps exploited for both classification and regression tasks.

To sum up, the contributions of this work are summarized as follows:

- We propose a novel and effective method RFN for multi-orientation object detection, which provides rotation-sensitive and rotation-invariant feature maps to meet the different requirements for regression and classification. The RFN can achieve great performance with only a slight increase in model complexity.
- To make the rotation-invariance more reliable, we adopt a metric to quantitatively evaluate it by adding a constrain item in the loss, yielding a promising result for multi-orientation object detection.
- The RFN shows good generalization on the scene classification in remote sensing images and object detection in natural images. Extended experiments on NWPU-RESISC45 and Pascal VOC 2007 benchmarks show comparable or even better performance compared with the state-of-the-art methods.
- The Encoder unit of RFN is partly inspired by SE block. The experiment shows that we successfully upgrade SE block applied in multi-orientation object detection, leading to better detection performance.

## 2. Related Work

**Object Detection.** Benefited from the successfully application of deep convolutional networks, general object detectors [4, 7, 8, 20, 23, 24, 25] lead to great improvement for object detection, and have been widely used in practice. Generally, R-CNN [8] firstly exploits deep ConvNets in object detection and shows dramatic improvement compared with traditional methods. Inspired from R-CNN, a lot of CNN-based detectors are proposed, such as [4, 7, 25], and achieve greater performance in speed and accuracy. Recently, faster one-stage detectors have drawn much attention, such as [19, 20, 23, 24], which incorporate classification and regression tasks into a single stage.

Most object detectors share the same feature maps between a classifier and regressor for classification and localization, respectively. But for multi-orientation object detection, there exists conflicting requirements between classification and regression tasks. Different from previous works, our method provides rotation-invariant feature maps for classification, and rotation-sensitive feature maps for regression. In this manner, we can meet the different requirements in multi-orientation object detection, especially for remote sensing images.

**Feature Representation**. Recent deep ConvNets [12, 15, 16, 33, 38] provide great feature representation ability, leading to cutting-edge performance in a wide range of tasks, such as image classification, language recognition and object detection. Inception modules [33] and VGGNets [32] demonstrate the powerful representation ability by exploiting the deep architecture. ResNets [12] add a skip connection from input to output, which can largely address the degradation problem. The ConvNets can achieve rotation-invariance with max pooling layer. Recently, carefully designed components, such as ORN [39] and RRD [17], can produce rotation-invariant feature maps, which can also enhance rotation-invariance for ConvNets.

However, previous works provide rotation-invariant features with only qualitative or intuitive analysis. In this paper, to make the rotation-invariant features more reliable, we adopt a metric to evaluate whether an operator is rotation-invariant. We add a constraint item in the loss, which is imposed on the rotation-invariant feature maps from RFN. Minimizing the loss can provide reliable rotation-invariance, yielding improvement in detection performance.

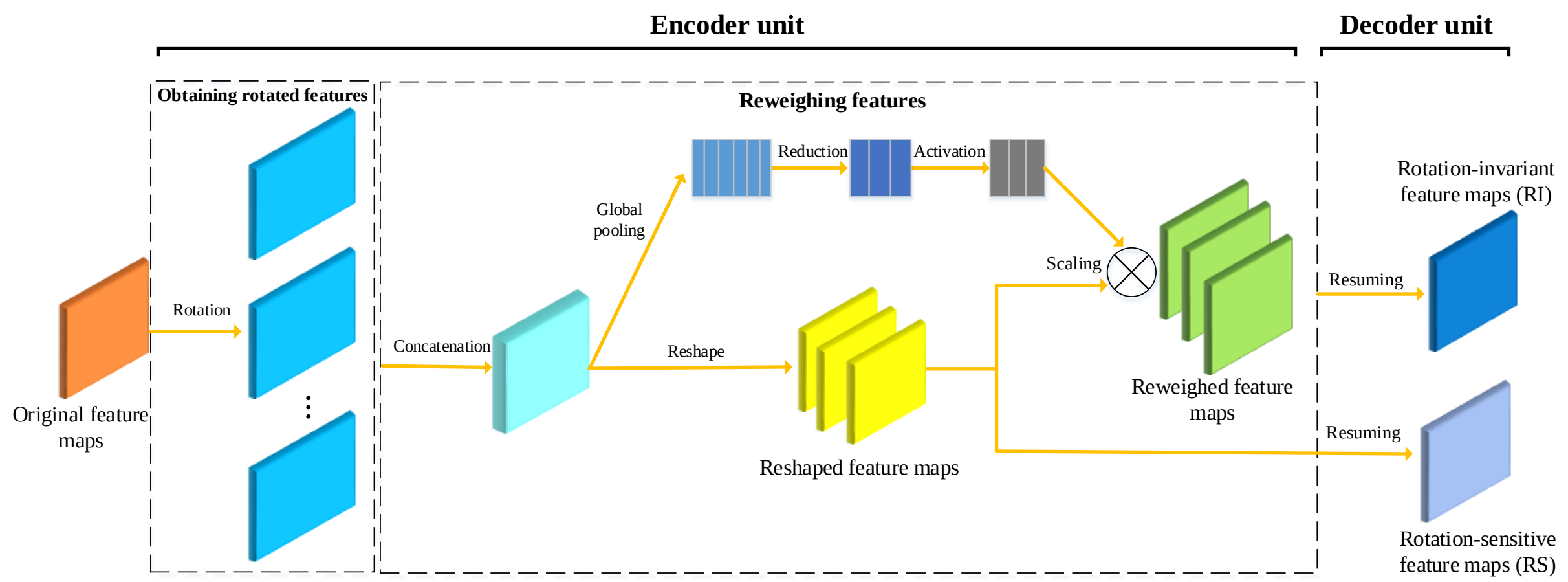

Figure 2. Overall architecture of the proposed RFN method.

# 3. The Proposed Model

## 3.1. Overview

The overall architecture is illustrated in Fig. 2. The whole process for object detection is based on an existing framework, i.e. Faster R-CNN. Original feature maps, produced from RoI pooling layer, are exploited as the input of RFN. RFN is a block extracting rotation-invariant and rotation-sensitive features maps. It is composed of two components: Encoder unit and Decoder unit. In the Encoder unit, the feature maps are rotated channel-wise with preset angles. The rotated feature maps are fed into two branches. The top branch produces feature weights for each angle, which is used to scale the reshaped feature maps produced from the down branch. The Decoder unit is exploited to produce rotation-invariant and rotation-sensitive feature maps. The following processes are commonly used in traditional object detection frameworks, namely, regression and classification. We will fully elaborate the Encoder unit, Decoder unit, localization of RFN and training strategy.

## 3.2. Encoder Unit

SENet [14] is a novel architecture designed for improving the representation ability of a network, which shows great improvement for existing frameworks. The Encoder unit is an upgradation of SE block, where two new functions are developed, i.e. generation and fusion for rotated features. Generally, the Encoder unit is composed of two processes: obtaining rotated features and reweighing features. In this unit, obtaining rotated features is firstly conducted, which makes the channels expanded. Then features are scaled angel-wise in the process of reweighing features.

**Obtaining rotated features.** Previous works try to obtain rotated features by rotating images in the preprocessing process [2] or using rotated convolutional filters [17, 39]. Here, a simply and easily implemented method is proposed. Assume $X \in R^{1\times W\times H\times C}$ denotes the original feature maps, and $f_\theta(\cdot)$ is a rotation operation, which rotates the input counterclockwise by angle $\theta$. The rotated feature maps can be formulated as below:

$$Y_i^\theta = f_\theta(X_i), \tag{1}$$

where $X_i \in R^{1\times W\times H\times 1}$ denotes the i-th channel of $X$. With the rotation operation, we can obtain the rotated feature maps $Y^\theta \in R^{1\times W\times H\times C}$ by rotating $X$ channel-wise. Note that we use a resize operation after rotation to keep the size.

Given $Y = \{Y^{\theta_0}, Y^{\theta_1}, \cdots, Y^{\theta_{n-1}}\}$, each element represents the original feature maps rotated by a specific angle. For each angle $\theta_k$, it is calculated as below:

$$\theta_k = k\frac{2\pi}{n}, \mathrm{k} = 0,1,\cdots,n-1. \tag{2}$$

Note that there are $n$ feature maps contained in $Y$. In order to obtain single feature maps, we define an operation named Concatenation, which concatenates the $n$ elements over the channel dimension. Finally, the final rotated feature maps can be calculated as follows:

$$M = g(Y) = g(Y^{\theta_0}, \cdots, Y^{\theta_{n-1}}), \tag{3}$$

where $g(\cdot)$ denotes the Concatenation operation, and $M$ represents the output with size $1\times W\times H\times(n\times C)$. It can be seen that the size of channel dimension is expanded by $n$ times.

**Reweighing features.** Different orientation information leads to different effects on the final results. Therefore, we should evaluate how the feature maps of different angles influence the results. An intuitive idea is to assign a weight to the feature maps of each angle. We apply this idea in the reweighing features process, which includes two branches, the top for generating weights and the down for assigning feature maps to each angle.

In the top branch, we firstly extract the global information by performing a Global pooling operation on rotated feature maps, which is simply taking advantage of all $W \times H$ values. In this paper, we investigate two Global pooling operations in section 4.3: global max pooling and global average pooling. Here, we just take global max pooling for example. Specifically, $M$ is fed into the global max pooling operation, and $G$ is generated from this operation. The c-th element of $G$ is calculated as follows:

$$G_c = F_{gmp}(M_c) = \max\{M_c(i,j) \mid \forall M_c(i,j) \in M_c\}, \quad (4)$$

where $F_{gmp}(\cdot)$ denotes the max pooling operation, and $M_c$ denotes the feature maps of the c-th channel. It can be inferred that the size of $G$ is $1 \times (n \times C)$.

To generate weights for feature maps of each angle, we use two operations. The Reduction operation is used to reduce the size of second dimension to $n$. The Activation operation is expected to map the $n$ values to the range (0, 1). A two-step reduction strategy is adopt. We exploit two linear layers to achieve this target, each of which is followed with an activation function. We also investigate how the reduction ratio of the first linear layer influences the performance in section 4.3. Assume $\omega \in R^{1\times n}$ represents the weights. It is calculated as follows:

$$\omega = F_{sigmoid}(F_{ReLU}(GW_1)W_2), \quad (5)$$

where $W_1$ and $W_2$ denote the parameters of the first and second linear layers, respectively, $F_{ReLU}$ represents the ReLU [21] function for providing non-linear ability, and $F_{sigmoid}$ denotes the sigmoid function for mapping values to the range (0, 1).

In the down branch, we simply use the reshape operation to obtain the feature maps of each angle. Assume that the reshaped feature maps $M' \in R^{n \times W \times H \times C}$ are produced from the reshape operation, the reweighed feature maps can be obtained by using the scaling operation as below:

$$S_i = F_{scale}(\omega, M') = \omega_i \times M'_i, \quad (6)$$

where $M'_i$ is correspond to the rotated feature maps with angle $\theta_i$, $F_{scale}$ represents the scaling operation, and $S_i$ denotes the reweighed result of $M'_i$.

### 3.3. Decoder Unit

For multi-orientation object detection, it is obvious that object localization is sensitive to object orientations, but object classification is rotation-invariant, as a specific object of arbitrary orientations should be classified to the same class. Therefore, there exists incompatible demands for regression and classification. One target of Decoder unit is to produce two types of feature maps to meet different demands for regression and classification, i.e. RI and RS, which are extended from the reweighed and reshaped

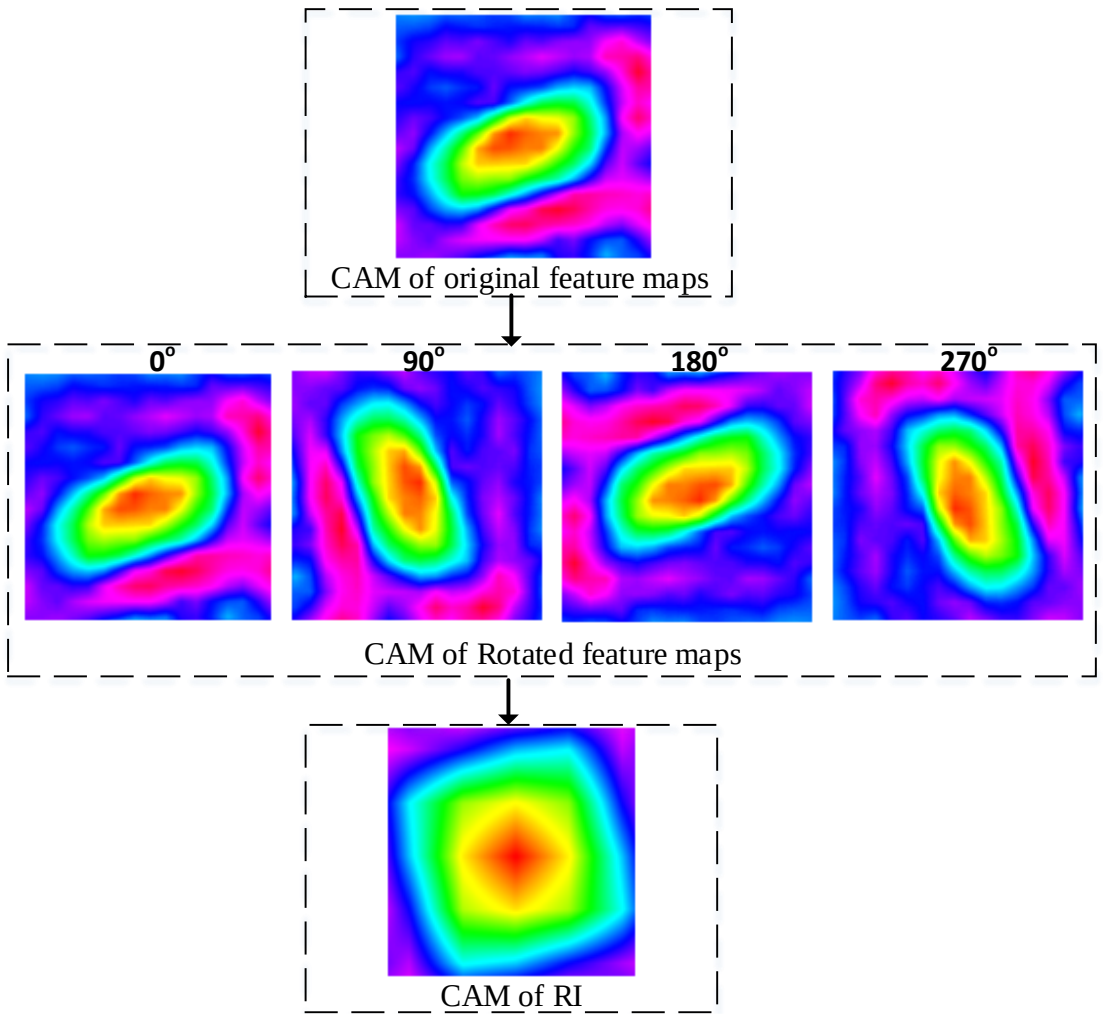

Figure 3. Visualization of CAM for original feature maps and rotation-invariant feature maps.

feature maps, respectively.

It can be inferred that the sizes of the reweighed and reshaped feature maps from Encoder unit are both $n \times W \times H \times C$, which are $n$ times of the original feature maps in the first dimension. Another target of Decoder unit is to reduce the reweighed and reshaped feature maps to the original size. There are two points to explain the design purpose: 1) The proposed RFN is used to further extract high-level features. Once the size of input and output feature maps are kept identical, the RFN can be used as an independent block embedded into an existing framework with minor modifications. 2) The training and testing phases will consume more time with the increasing angle number. So reducing to the original size can improve the efficiency of the whole model. Besides, it also helps to reduce redundant features.

To achieve the above targets, a Resuming operator is expected to handle over the first dimension of the reweighed and reshaped feature maps. We investigate two types of Resuming operators in section 4.3: max and sum. Here, we just describe how to obtain RI with the max operation. Assume $S = \{S_0, S_2, \cdots, S_{n-1}\}$ and $O \in R^{1\times H\times W\times C}$ denotes RI, which can be calculated as follows:

$$O = \max\{S_i \mid \forall S_i \in S\}. \quad (7)$$

To demonstrate how the feature maps change in RFN, we visualize the class activation maps (CAM) for the RI. As illustrated in Fig. 3, the original feature maps are rotated by 4 angles in the RFN. After a series of operations performed on the original feature maps, the produced RI are more robust and insensitive to orientations. Compared with the original feature maps, the RI takes more comprehensive information for final recognition, largely eliminating the influence of orientations.

### 3.4. Localization of RFN

Region-based systems [4, 9] for object detection have been widely used, and possessed leading performance on several benchmarks [6, 18, 27]. Note that the original feature maps are attached to a specific region, so it represents the feature maps of a candidate region, which we call region-based feature maps. Therefore, the proposed RFN (shown in Fig. 4(a)) is a region-based method, as it takes region-based feature maps as input. A global RFN (shown in Fig. 4(b)) is also an alternative. Different from the region-based architecture, global RFN takes features maps with global information as input, which is produced from previous convolutional layers over the whole image. Compared with global RFN, region-based RFN takes advantage of more accurate localization information to generate rotation-sensitive features of a specific object. The two types of RFN are investigated in section 4.3.

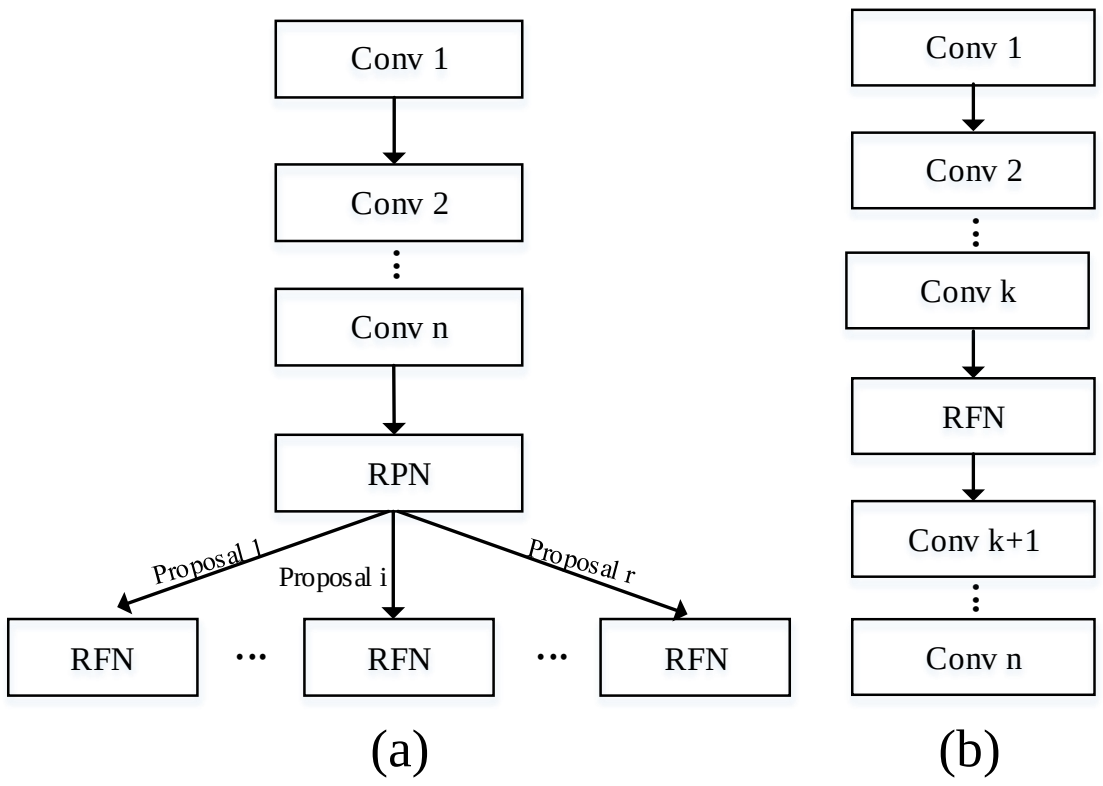

Figure 4. Type of RFN. (a): region-based RFN; (b): global RFN.

### 3.5. Training

The RS are extracted from rotated features with rich orientations, which is implemented by using the Resuming operator to select the most representative features. Therefore, the RS are highly related to orientation. We can intuitively regard RS as rotation-sensitive. But the problem is how to promise that the RI is rotation-invariant.

As noted in [5], an operator $G$ is Locally Rotation Invariant (LRI) as long as the response map at $y_0$ is not changed by the rotation of the input image around $y_0$. Specifically, for any $y_0 \in R^3$ , any rotation operator $R$, and any input image $I$, it follows

$$(G(I))_{y_0} = (G(R(I)))_{y_0}\ , \tag{8}$$

where $R(\cdot)$ and $\mathrm{G}(\cdot)$ are performed on an image, $(\cdot)_{y_0}$ denotes the element at position $y_0$ in an image. Inspired from Eq. (8), we present a metric to minimize the difference between the RIs before and after rotating the original feature maps. We define a loss to describe this metric as follows:

$$L_{ri} = \frac{1}{2N(n-1)}\sum_{i=0}^{N-1}\sum_{j=1}^{n-1} k(\mathrm{y}_i^0, y_i^{\theta_j})\ , \tag{9}$$

where $k(\cdot,\cdot)$ is a Gaussian radial basis kernel function to calculate the difference between two feature maps, $y_0$ and $y_{\theta_i}$ denote the produced RIs from RFN by taking the original feature maps with no rotation and rotation by angle $\theta_i$ as input, respectively. $N$ is the batch size. The kernel function aims to map the features into the same feature space, which can bridge the distribution difference caused by the forward propagation in the neural network. Note that the RIs should be normalized before calculating $L_{ri}$ . In practice, since calculating $L_{ri}$ costs a lot of time, we just task the following as a substitute

$$L_{ri}^{*} = \frac{1}{N}\sum_{i=0}^{N-1} k(\mathrm{y}_i^0, \overline{y_i^{\theta}})\ , \tag{10}$$

where $\overline{y_i^{\theta}}$ denotes the average of $y_i^{\theta_j}$ for $j = 1, \cdots, n-1$. In the training phase, the output boxes are matched to the ground-truth boxes according to the scheme described in [7]. Similar to the base framework, we adopt multitask loss function. The total loss can be represented as follows:

$$L = \frac{1}{N_{cls}}\sum_{i=1}^{N_{cls}} L_{cls}(p_i, p_i^*) + \lambda_{reg}\frac{1}{N_{reg}}\sum_{j=1}^{N_{reg}} L_{reg}(t_j, t_j^*) + \lambda_{ri} L_{ri}^*\ , \tag{11}$$

where $p_i^*$ denotes the ground-truth labels, $p_i$ is the predicted probability, $t_j$ represents a vector representing the parameterized coordinates of the predicted bounding box, and $t_j^*$ indicates the ground-truth offsets of a generated box. In all experiments, the regularization items $\lambda_{reg}$ and $\lambda_{ri}$ are set to 0.2 and 0.5 for quick convergence. The smooth L1 loss is used for $L_{reg}$ , and log loss over two classes is adopt for classification loss $L_{cls}$ .

## 4. Experiments

### 4.1. Datasets Description

**NWPU VHR-10** contains 800 very-high-resolution (VHR) remote sensing images with a wide range of spatial resolutions. All the images are taken from Google Earth with spatial resolutions from 0.5 m to 2.0 m and Vaihingen with a spatial resolution 0.08m. It is a widely used dataset with 10 classes of objects, such as airplane, ship, storage tank, baseball diamond et al.

**RSOD** comes from Google Earth and Tianditu with 4 classes: oil tank, aircraft, overpass, and playground. Compared to NWPU VHR-10, this dataset has also a wide range of spatial resolutions, varying from 0.3-3 m. RSOD contains 2326 images annotated by experts.

**Pascal VOC 2007** is a natural image dataset, which is commonly used to evaluate a detector's performance. There are 20 object class in the dataset, such as bird, cat, dog and boat. The dataset is just used for evaluating the generalization of RFN on object detection in natural images.

**NWPU-RESISC45** is a large scale dataset consisting of 31500 images. Images can be divided into 45 scenes, such as airport, beach, forest, harbor, lake, island et al. each scene is composed of 700 images with a fixed size of 256× 256 pixels. Most of the images have the spatial resolution varying from 0.2 m to 30 m, except for images in the scenes of island, lake, mountain, and snowberg with quite lower spatial resolutions. This dataset is used for the task of scene classification to further validate the efficiency of the proposed RFN in the related task.

## 4.2. Setup and Implementation Details

The input images are firstly resized to 800 × 800 and then randomly cropped to 640 × 640. We use randomly horizontal flipping as the only data augmentation method during training phase. We exploit two deep ConvNets, VGG-16 and ResNet-101, as the backbone network, which are pre-trained on the ImageNet dataset. The other layers are initialized with truncated normal method.

We use SGD as the optimizer. The weight decay and momentum are 0.0005 and 0.9, respectively. All the networks are trained for 30 epochs. The initial learning rate is 0.001 with a decay rate 0.95 every 10 epochs. The IoU threshold is set to 0.5. A GTX 1080TI is used for training and testing. Our experiment is implemented with the deep learning framework, pytorch.

| Item | Setting |
|---|---|
| Reduction ratio | 8 |
| Number of rotation angle | 4 |
| Type of RFN | Region-based |
| Type of Global pooling operation | Global max pooling |
| Type of Resuming operation | Sum |

Table 1. The default setting in the ablation study.

## 4.3. Ablation Study

To show how the components of RFN influence the performance, we conduct several experiments on the NWPU VHR-10 dataset. We use ResNet-101 as the backbone architecture. Without explicit statement, we use the default setting described in Table 1. The experiment conducted under the default setting is used as the baseline. All the variants of the RFN are generated by changing the default setting. We discuss the RFN in five aspects: reduction ratio, number of rotation angle, type of RFN, type of Global pooling operation and type of Resuming operation.

| | Method | mAP (%) | Parameters (M) |
|---|---|---|---|
| - | Original | 88.3 | **52.2** |
| Default setting | Baseline | **93.7** | 54.3 |
| Reduction ratio | R0 | 90.0 | **52.2** |
| | R4 | 90.7 | 56.4 |
| | R16 | 89.4 | 53.2 |
| | R32 | 91.9 | 52.7 |
| Number of rotation angle | A2 | 86.7 | 52.7 |
| | A6 | 86.1 | 56.9 |
| | A8 | 84.5 | 60.6 |
| Type of RFN | Global 1 | 91.0 | 52.2 |
| | Global 2 | 90.8 | 52.3 |
| | Global 3 | 89.9 | 52.7 |
| Type of global pooling operation | Global average pooling | 91.3 | 54.3 |
| Type of resuming operation | Max | 91.8 | 54.3 |

Table 2. The results in ablation study on NWPU VHR-10 dataset. Original represents Faster R-CNN with backbone ResNet-101.

**Reduction ratio**. Reduction ratio is a hyperparameter as mentioned in the process of reweighing features. We investigate how reduction ratio influences the result. The result is shown in Table 2. 4 different reduction ratios are conducted for comparison in this experiment. R4, R8 (baseline), R16, R32 denote reduction ratio of 4, 8, 16, 32, respectively. Moreover, in order to show how important the reduction operation is to the result, we remove the reduction operation in order to compare with other 4 reduction ratios, which is called R0. As a result, R8 achieves the best performance and outperforms R0 by 3.7 %. It can be inferred that the reduction operation has a positive effect on the result. Noted that the performance does not always rise with the reduction ratio increasing.

**Number of rotation angle**. Feature maps rotated by different angles may have different influence on the performance. We try 2, 4, 6, 8 angles to rotate feature maps, which are represented by A2, A4 (baseline), A6, A8, respectively. The result in Table 2 shows that A4 performs better than others. However, the model has lower complexity under A2. To make a balance between the performance and complexity, we recommend that the number of rotation angle is set to 4.

**Type of RFN.** Here we discuss two types of RFN: region-based and global RFN. For region-based RFN (baseline), the RFN is placed after RoI pooling, taking feature maps of each proposal as input. For global RFN, we investigate three kinds of global architecture. Resnet-101 has 4 cascaded residual layers, and only the first three residual layers are used for feature extraction in the experiment. We place RFN after each of the first three residual layers, forming three kinds of global architecture, i.e. Globa1, Global 2, and Global 3. The result in Table 2 indicates that the region-based RFN has a better ability in feature representation. A more accurate localization information generated from region-based RFN can explained for the result.

**Type of Global pooling operation**. Two types of global operation are investigated: global max pooling (baseline) and global average pooling. Global pooling operation aims to extract the most representative information over each channel of feature maps. From the result, global max pooling achieves improvement by about 2.4%. Therefore, global max pooling is recommended as the Global pooling operation.

| Method | plane | ship | storage tank | baseball diamond | tennis court | basketball court | ground track field | harbor | bridge | vehicle | mAP |
|---|---|---|---|---|---|---|---|---|---|---|---|
| RICNN | 88.4 | 77.3 | 85.3 | 88.1 | 40.8 | 58.5 | 86.7 | 68.6 | 61.5 | 71.1 | 72.6 |
| SSD | 90.4 | 60.9 | 79.8 | 89.9 | 82.6 | 80.6 | 98.3 | 73.4 | 76.7 | 52.1 | 78.5 |
| Faster R-CNN (VGG-16) | 99.7 | 88.1 | 39.7 | 90.9 | 80.3 | 97.0 | 99.7 | 99.0 | 70.0 | 75.6 | 84.0 |
| Faster R-CNN (ResNet-101) | 90.9 | 88.6 | 41.3 | 90.9 | 88.1 | 90.4 | 99.7 | 88.9 | 90.6 | 74.4 | 88.3 |
| R-FCN | 81.7 | 80.6 | 66.2 | 90.3 | 80.2 | 69.7 | 89.8 | 78.6 | 47.8 | 78.3 | 76.3 |
| Deformable R-FCN | 87.3 | 81.4 | 63.6 | 90.4 | 81.6 | 74.1 | 90.3 | 75.3 | 71.4 | 75.5 | 79.1 |
| $R^2$CNN++ | **100.0** | 89.4 | **97.2** | 97.0 | 83.2 | 87.5 | 99.2 | 99.4 | 74.5 | 90.1 | 91.8 |
| Ours (VGG-16) | 90.9 | 88.0 | 59.0 | **99.9** | 89.9 | 99.4 | **100.0** | 99.2 | 89.5 | 88.8 | 90.5 |
| Ours* (ResNet-101) | 99.8 | 86.4 | 76.5 | 99.8 | 90.4 | **100.0** | 99.7 | **100.0** | 80.8 | 86.4 | 92.0 |
| Ours (ResNet-101) | 98.0 | **89.7** | 83.6 | 98.2 | **93.3** | **100.0** | 96.7 | **100.0** | 85.1 | 92.4 | 93.7 |
| Ours (soft-NMS) | 96.4 | 88.7 | 75.5 | 99.3 | 89.7 | **100.0** | **100.0** | **100.0** | **99.0** | **92.9** | **94.2** |

Table 3. Comparison results (%) on NWPU VHR-10 dataset by using VGG-16 and ResNet-101. Ours* (ResNet-101) denotes $L_{ri}^{*}$ is not added in the total loss.

| Method | aero | bike | bird | boat | bottle | bus | car | cat | chair | cow | table | dog | horse | mbike | persn | plant | sheep | sofa | train | tv | mAP |
|---|---|---|---|---|---|---|---|---|---|---|---|---|---|---|---|---|---|---|---|---|---|
| SSD | 75.1 | 81.4 | 69.8 | 60.8 | 46.3 | **82.6** | **84.7** | 84.1 | 48.5 | 75.0 | 67.4 | 82.3 | **83.9** | **79.4** | 76.6 | 44.9 | 69.9 | 69.1 | 78.1 | 71.8 | 71.6 |
| OHEM | 71.2 | 78.3 | 69.2 | 57.9 | 46.5 | 81.8 | 79.1 | 83.2 | 47.9 | 76.2 | 68.9 | 83.2 | 80.8 | 75.8 | 72.7 | 39.9 | 67.5 | 66.2 | 75.6 | 75.9 | 69.9 |
| Faster R-CNN | 70.0 | 80.6 | 70.1 | 57.3 | 49.9 | 78.2 | 80.4 | 82.0 | 52.2 | 75.3 | 67.2 | 80.3 | 79.8 | 75.0 | 76.3 | 39.1 | 68.3 | 67.3 | **81.1** | 67.6 | 69.9 |
| RIFD-CNN | **78.9** | 82.5 | 69.6 | 54.2 | 49.7 | 78.3 | 82.0 | 83.4 | 51.1 | 76.0 | 69.0 | 82.2 | 80.7 | 77.2 | 73.1 | 42.6 | **70.3** | 70.4 | 74.2 | 74.1 | 71.0 |
| CC-Net | 78.3 | 79.4 | 69.1 | 63.5 | **53.2** | 82.1 | 79.7 | 86.3 | 56.0 | 75.6 | **72.3** | **83.4** | 79.0 | 76.3 | 76.4 | 43.1 | 67.6 | **71.8** | 77.3 | **76.6** | 72.4 |
| Ours | 76.9 | **83.7** | **70.6** | **65.1** | 49.4 | 79.2 | 81.8 | **86.6** | **58.9** | **77.7** | 68.5 | 78.9 | 83.5 | 78.0 | **78.5** | **48.9** | 68.0 | 66.8 | 78.8 | 71.6 | **72.6** |

Table 5. Comparison results (%) with the state-of-the-art methods on Pascal VOC 2007 dataset by using VGG-16.

**Type of Resuming operation**. As pointed in section 3.3, the Resuming operator has two choices: max and sum (baseline). From Table 2, the comparison between max and sum demonstrates that sum has a better ability in extracting and fusing high-level features. This is because the max operation only preserve the maximum element over the first dimension, resulting in missing a lot of information. On the contrary, the sum operation exploits all the information to produce rotation-invariant and rotation-sensitive feature maps.

| Method | aircraft | oil tank | overpass | playground | mAP |
|---|---|---|---|---|---|
| R-P-Faster R-CNN | 70.8 | 90.2 | 78.7 | 98.1 | 84.5 |
| Deformable R-FCN | 71.5 | 90.3 | 81.5 | 99.5 | 85.7 |
| Deformable R-FCN and arcNMS | 71.9 | 90.4 | 89.6 | 99.9 | 87.9 |
| Faster R-CNN (VGG-16) | 71.3 | 90.7 | 90.9 | 99.7 | 88.1 |
| Faster R-CNN (ResNet-101) | 71.9 | 90.9 | **100.0** | **100.0** | 90.7 |
| SE block (VGG-16) | **80.5** | 90.9 | 90.9 | **100.0** | 90.6 |
| SE block (ResNet-101) | 77.4 | 90.9 | **100.0** | 99.1 | 91.8 |
| Ours (VGG-16) | 80.1 | **99.4** | **100.0** | **100.0** | **94.9** |
| Ours (ResNet-101) | 79.1 | 90.5 | **100.0** | 99.7 | 92.3 |

Table 4. Detection results (%) on RSOD dataset using VGG-16 and ResNet-101 as backbone architecture.

## 4.4. Comparison with State-of-the-art

**NWPU VHR-10.** We compare the proposed RFN method with several state-of-the-art deep learning-based ones as follows: RICNN [3], SSD [20], Faster R-CNN [25], Deformable Faster R-CNN [26], R-FCN [4], Deformable R-FCN [35], $R^2$CNN++ [37]. For a fair comparison, we adopt the default settings as described in original papers. VGG-16 and ResNet-101 are adopt as backbone architectures. Faster R-CNN acts as the baseline method.

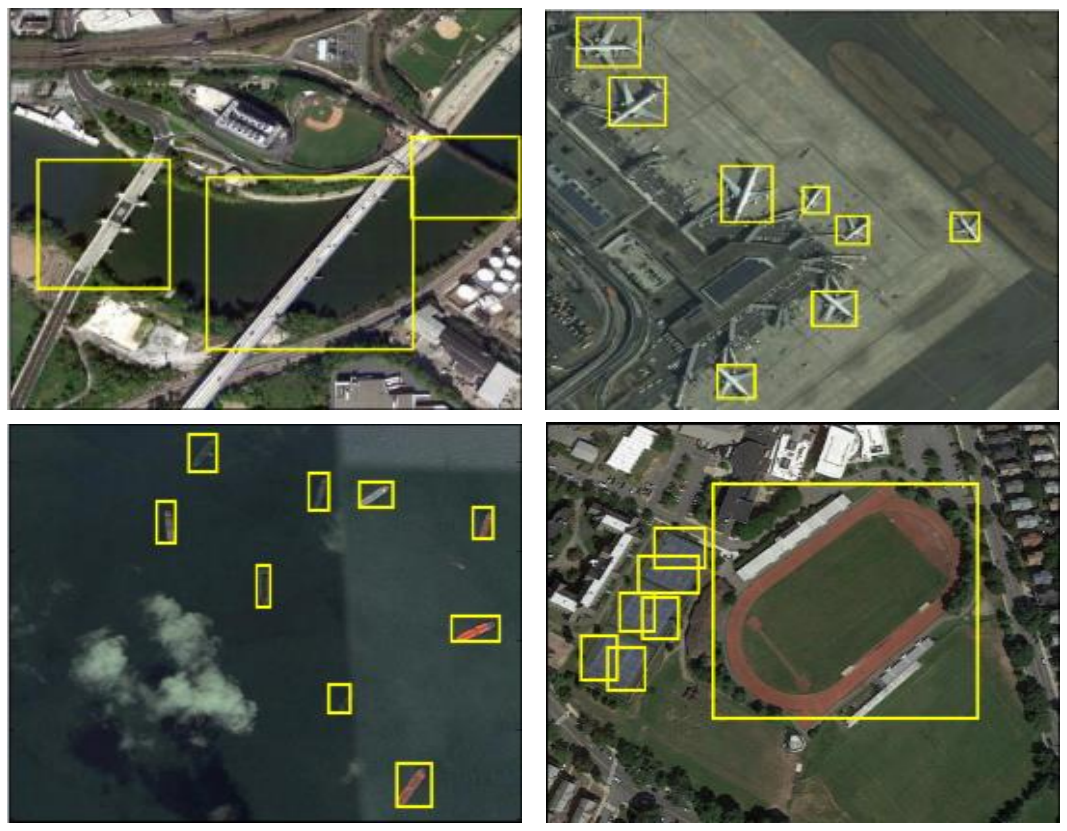

Figure 5. Some detection results on NUPU VHR-10 dataset.

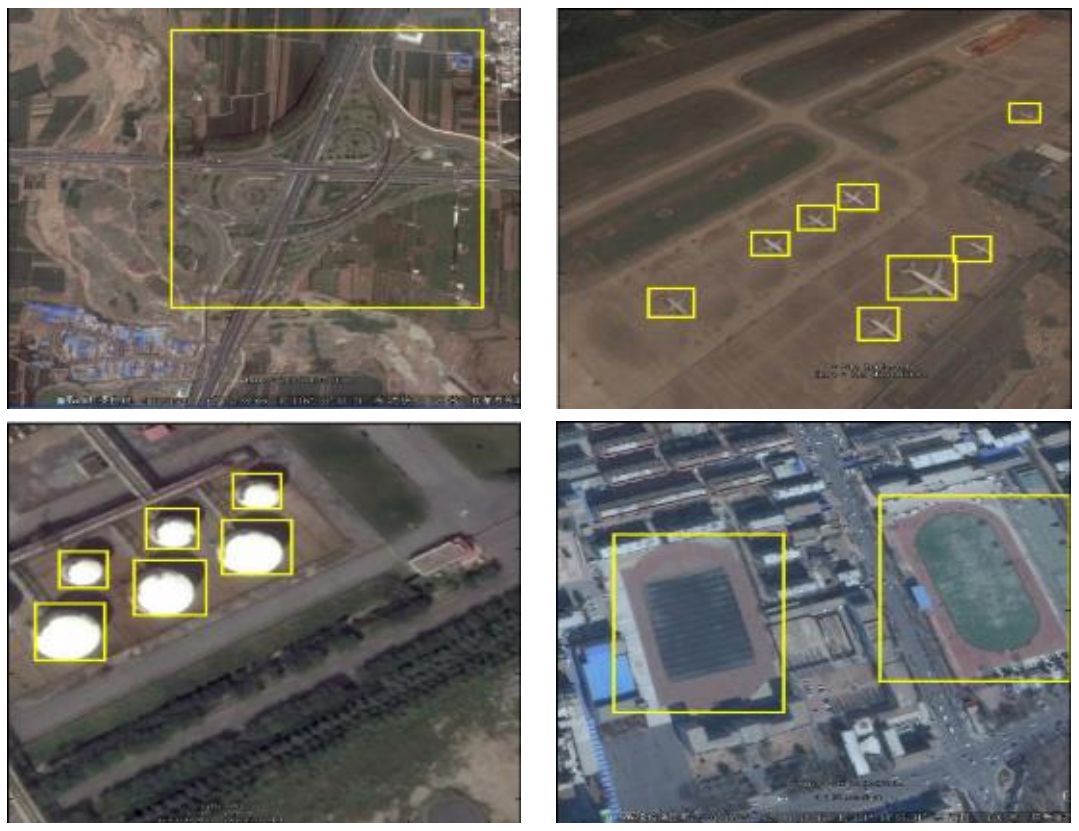

Figure 6. Some detection results on RSOD dataset using VGG-16 and ResNet-101.

The results on NWPU VHR-10 dataset are reported in Table 3. Our method outperforms the baseline method by about 5.4% while using ResNet-101 as backbone. Moreover, the results in Table 2 shows that our method improves performance with only a slight increase in model complexity. If we replace ResNet-101 with VGG-16, our method has a significant improvement over the base framework by about 6.5%. Compared with the recently proposed $R^2$CNN++, our method also improves the detection performance by about 1.9%. Surprisingly, on the classes 'tennis court', 'basketball court' and 'bridge', our method performs better than $R^2$CNN++ by at least 10%. Moreover, if we only use soft-NMS in the test phase without modification in the training phase, there is a direct improvement by 0.5%. We further evaluate how the evaluation metric for rotation-invariance influence the performance. As a result, this metric can improve the performance by 1.7%. Some visualization results are shown in Fig. 5. It can be seen that our method can detect objects with arbitrary orientations, especially for ship and plane.

**RSOD.** We evaluate the proposed RFN on RSOD dataset and compare with the state-of-the-art detectors: R-P-Faster R-CNN [11], Deformable R-FCN [35], Deformable ConvNet with arcNMS [36], Faster R-CNN [25]. We also use VGG-16 and ResNet-101 as the backbone architecture. The results are shown in Table 4. Compared with the baseline method, our method demonstrates greater performance with both VGG-16 and ResNet-101. We also make a comparison with SE block by replacing RFN with it. As a result, it can be seen that we successfully upgrade SE block to form the proposed RFN, achieving state-of-the-art performance. Some detection results are displayed in Fig. 6.

**Pascal VOC 2007.** Pascal VOC 2007 is a widely used dataset containing lots of natural images. Although the RFN is mainly designed for remote sensing images, there still exists slight rotation invariance in natural images. In order to further evaluate the detection performance, our method is trained and evaluated on VOC07 trainval and test datasets, respectively. We compare with several state-of-the-art methods: SSD [20], OHEM [31], Faster R-CNN [25], RIFD-CNN [2] and CC-Net [22]. We adopt VGG-16 as backbone for a fair comparison. Faster R-CNN is our baseline method. The comparison results are reported in Table 5. Compared with the baseline method, the detection performance can be improved by about 2.7%. RIFD-CNN is also mainly designed for the detection task where object orientations vary a lot. Our method outperforms it by about 1.6%. Our method also shows comparative results compared with other state-of-the-art methods. For better demonstrating the detection performance, we carefully choose some examples whose orientations changes relatively largely. The visualization results are displayed in Fig. 7. It can be observed that our method can well handle the detection task where objects in natural images are slight multi-orientation.

## 4.5. Classification Results

To evaluate the generalization ability of RFN, we evaluate it on NWPU-RESISC45 dataset. Three convolutional neural networks are served as the baseline models, i.e. AlexNet [16], GoogleNet [33] and VGG-16

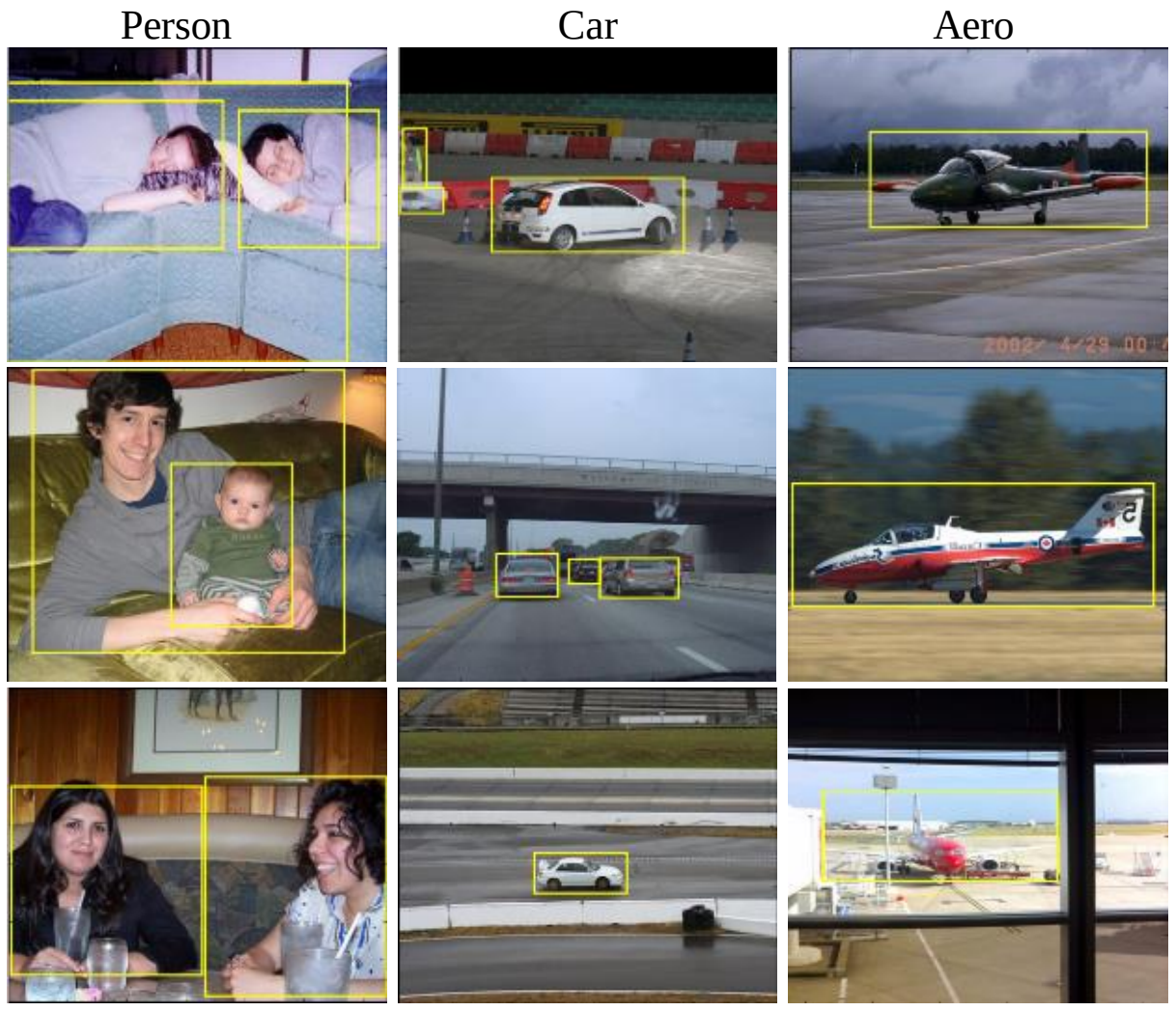

Figure 7. Visualization of some results on Pascal VOC 2007 dataset.

| Method | Training ratio | |
|---|---|---|
| | 10% | 20% |
| GoogleNet | 76.2 | 78.5 |
| AlexNet | 76.7 | 79.9 |
| VGG-16 | 76.5 | 79.8 |
| GoogleNet-RFN | 80.9 | 84.2 |
| AlexNet-RFN | 81.4 | 84.4 |
| VGG-RFN | **82.8** | **88.9** |

Table 6. The classification accuracies (%) compared with three convolutional neural networks on NWPU-RESISC45 dataset.

[32]. The RFN is inserted between the last Conv layer and classifier, forming AlexNet-RFN, GoogleNet-RFN and VGG-RFN. To adapt the classification task, we only connect the RI with the classifier. For a fair comparison, the linear SVM (C=1) is used as classifier for all methods. To make a comprehensive evaluation, we use two training set ratios, i.e. 10% and 20%. The accuracy is adopt as the evaluation metric.

The comparison results are described in Table 6. While using the VGG-16, the proposed RFN leads to dramatic improvement by about 9% for the large training ratio. Our method performs better than all baseline models by at least 4.5%. It is an interesting trend that our method can yield a larger improvement when more training samples are used.

# 5. Conclusion

We propose RFN, a novel and effective method designed for multi-orientation object detection, which produces rotation-invariant and rotation-sensitive feature maps for classification and regression, respectively. The RFN can be integrated into an existing framework, achieving great performance with only a slight increase in model complexity. Moreover it shows great generalization ability on scene classification in remote sensing images and object detection in natural images. Future works will focus on multi-orientation object detection with oriented boxes.